\documentclass[11pt,a4paper]{article}
\usepackage{arabtex}
\usepackage[hyperref]{acl2017}
\usepackage{times}
\usepackage{amssymb}
\usepackage{amsmath}

\usepackage{latexsym}
\usepackage{url}
\usepackage{amssymb}
\usepackage{algorithmic}
\usepackage[ruled,vlined]{algorithm2e}
\usepackage{tikz}
\usepackage{verbatim}
\usetikzlibrary{arrows,shapes}
\usepackage{multirow}
\usepackage{amsmath}
\usepackage{booktabs}
\usepackage{enumerate}
\usepackage{graphicx}
\usepackage{mathtools}
\DeclareMathAlphabet{\pazocal}{OMS}{zplm}{m}{n}

\usepackage{microtype}
\usepackage{arabtex}
\usepackage{utf8}
\setcode{utf8}

\usepackage[colorinlistoftodos]{todonotes}
\DeclareMathAlphabet{\pazocal}{OMS}{zplm}{m}{n}

\DeclareMathOperator*{\argmin}{\arg\!\min}

\aclfinalcopy 

\newcommand{\Ni}{({\em i})~}
\newcommand{\Nii}{({\em ii})~}
\newcommand{\Niii}{({\em iii})~}

\newcommand{\Na}{({\em a})~}
\newcommand{\Nb}{({\em b})~}
\newcommand{\Nc}{({\em c})~}
\newcommand{\Nd}{({\em d})~}
\newcommand{\Ne}{({\em e})~}
\newcommand{\Nf}{({\em f})~}

\newcommand{\Ds}{\pazocal{D}}
\newcommand{\Rs}{\pazocal{R}}
\DeclareMathOperator*{\sigm}{sigm}
\newcommand{\Ls}{\pazocal{L}}

\title{Cross-language Learning with Adversarial Neural Networks:\\
Application to Community Question Answering}

\author{Shafiq Joty, Preslav Nakov, Llu\'is M\`arquez \and Israa Jaradat\\
ALT Research Group\\
Qatar Computing Research Institute, HBKU \\
  {\tt \{sjoty, pnakov, lmarquez, ijaradat\}@hbku.edu.qa}}

\date{}

\begin{document}

\maketitle

\begin{abstract}

We address the problem of cross-language adaptation for question-question similarity reranking in community question answering, with the objective to port a system trained on one input language to another input language given labeled training data for the first language and only unlabeled data for the second language.
In particular, we propose to use adversarial training of neural networks to learn high-level features that are discriminative for the main learning task, and at the same time are invariant across the input languages. The evaluation results show sizable improvements for our cross-language adversarial neural network (CLANN) model over a strong non-adversarial system.





\end{abstract}

\section{Introduction}

Developing natural language processing (NLP) systems that can work indistinctly 
with different input languages is a challenging task; yet, such a setup is useful for many real-world applications. One expensive solution is to annotate data for each input language and then to train a separate system for each one. 
Another option, which can be also costly, is to translate the input, e.g., using machine translation (MT), and then to work monolingually in the target language~\cite{Hartrumpf:08,Lin:10,ture-boschee:2016:EMNLP2016}.
However, the machine-translated text can be of low quality, 
might lose some input signal, e.g., it can alter sentiment \cite{Mohammad:2016:TAS},
or may not be really needed \cite{Bouma:08,pouranbenveyseh:2016:TextGraphs-10}.
Using a unified cross-language representation of the input is a third, less costly option, which allows any combination of input languages during both training and testing.

\noindent In this paper, we take this last approach, i.e., combining languages during both training and testing, and we study the problem of question-question similarity reranking in community Question Answering (cQA), when the input question can be either in English or in Arabic, and the questions it is compared to are always in English. We start with a simple language-independent representation based on cross-language word embeddings, which we input into a feed-forward multilayer neural network to classify pairs of questions, (English, English) or (Arabic, English), regarding their similarity. 

Furthermore, we explore the question of 
whether \emph{adversarial} training can be used to improve the performance of the network when we have some \emph{unlabeled} examples in the target language. In particular, we adapt the Domain Adversarial Neural Network model from~\cite{Ganin:2016:DTN:2946645.2946704}, which was originally used for domain adaptation, to our cross-language setting. To the best of our knowledge, this is novel for cross-language question-question similarity reranking, as well as for natural language processing (NLP) in general; moreover, we are not aware of any previous work on \emph{cross-language} question reranking for community Question Answering.

In our setup, the basic task-solving 
network is paired with another network that shares the internal representation of the input and tries to decide whether the input example comes from the source (English) or from the target (Arabic) language. The training of this language discriminator network is \emph{adversarial}
with respect to the shared layers by using gradient reversal during backpropagation, which makes the training to \emph{maximize} the loss of the discriminator rather than to minimize it. The main idea is to learn a high-level abstract representation that is discriminative for the main classification task, but is invariant across the input languages. 

\noindent We apply this method to an extension of the SemEval-2016 Task 3, subtask B benchmark dataset for question-question similarity reranking~\cite{nakov-EtAl:2016:SemEval}. In particular, we hired professional translators  to translate the original English questions to Arabic, and we further collected additional unlabeled questions in English, which we also got translated into Arabic. We show that using the unlabeled data for adversarial training allows us to improve the results by a sizable margin in both directions, i.e., when training on English and adapting the system with the Arabic unlabeled data, and vice versa. Moreover, the resulting performance is comparable to the best monolingual English systems at SemEval.
We also compare our unsupervised model to a semi-supervised model, where we have some labeled data for the target language.





The remainder of this paper is organized as follows:
Section~\ref{sec:related} discusses some related work.
Section~\ref{sec:model} introduces our model for adversarial training for cross-language problems.
Section~\ref{sec:setting} describes the experimental setup.
Section~\ref{sec:eval} presents the evaluation results.
Finally, Section~\ref{sec:conclusion} concludes and points to possible directions for future work.

\section{Related Work}
\label{sec:related}

Below we discuss three relevant research lines: \Na adversarial training, \Nb question-question similarity, and \Nc cross-language learning.

\emph{Adversarial training} of neural networks has shown a big impact recently, especially in areas such as computer vision, where generative unsupervised models have proved capable of synthesizing new images
\cite{Goodfellow_14_GAN,RadfordMC15,MakhzaniSJG15}. One crucial challenge in adversarial training is to find the right balance between the two components: the generator and the adversarial discriminator. Thus, several methods have been proposed recently to stabilize training \cite{MetzPPS16,ArjovskyCB17}. Adversarial training has also been successful in training predictive models. More relevant to our work is the work of \newcite{Ganin:2016:DTN:2946645.2946704}, who proposed domain adversarial neural networks (DANN) to learn discriminative but 
at the same time domain-invariant representations, with domain adaptation as a target. Here, we use adversarial training to learn task-specific representations in a \emph{cross-language} setting, which is novel for this task, to the best of our knowledge. 


\noindent \emph{Question-question similarity} was part of Task 3 on cQA at SemEval-2016/2017 \cite{nakov-EtAl:2016:SemEval,SemEval-2017:task3}; there was also a similar subtask as part of SemEval-2016 Task 1 on Semantic Textual Similarity \cite{agirre-EtAl:2016:SemEval1}. Question-question similarity is an important problem with application to question recommendation, question duplicate detection, community question answering, and question answering in general.
Typically, it has been addressed using a variety of textual similarity measures. Some work has paid attention to modeling the question topic, which can be done explicitly, e.g., using a graph of topic terms \cite{Cao:2008:RQU:1367497.1367509}, or implicitly, e.g., using LDA-based topic language model that matches the questions not only at the term level but also at the topic level \cite{zhang2014question}.
Another important aspect is syntactic structure, e.g., \citet{wang2009syntactic} proposed a retrieval model for finding similar questions based on the similarity of syntactic trees, and \citet{DaSanMartino:CIKM:2016} used syntactic kernels. Yet another emerging approach is to use neural networks, e.g., \citet{dossantos-EtAl:2015:ACL-IJCNLP} used convolutional neural networks (CNNs), \citet{Romeo:2016coling} used long short-term memory (LSTMs) networks with neural attention to select the important part when comparing two questions, and \citet{LeiJBJTMM16} used a combined recurrent--convolutional
model to map questions to continuous semantic representations.
Finally, translation models have been popular for question-question similarity \cite{Jeon:2005:FSQ:1099554.1099572,zhou2011phrase}.
Unlike that work, here we are interested in \emph{cross-language adaptation} for question-question similarity reranking.
The problem was studied in \cite{Giovanni2017_sigir} using cross-language kernels and deep neural networks; however, they used no adversarial training.

\emph{Cross-language Question Answering} was the topic of several challenges, e.g.,
at CLEF 2008~\cite{clef/FornerPAAFMOPRSSS08}, at NTCIR-8
\cite{mitamura:10}, and at BOLT
\cite{Soboroff:2016}. 
Cross-language QA methods typically use machine translation directly or adapt MT models to the QA setting \cite{Berger:2000:BLC:345508.345576,Echihabi:2003:NAQ:1075096.1075099,Soricut:2006:AQA:1127331.1127342,riezler-EtAl:2007:ACLMain,Hartrumpf:08,Lin:10,Surdeanu:2011:LRA:2000517.2000520,ture-boschee:2016:EMNLP2016}. They can also map terms across languages using Wikipedia links or BabelNet~\cite{Bouma:08,pouranbenveyseh:2016:TextGraphs-10}.
However, \emph{adversarial training} has not been tried in that setting.





\begin{figure}[tb!]
{\small
\begin{description}\setlength\itemsep{1pt}

\item[$q$:] \emph{give tips? did you do with it; if the answer is yes, then what the magnitude of what you avoid it? In our country, we leave a 15-20 percent.}\\
\vspace*{-3mm}

\item[$q'_1$] Tipping in Qatar. Is Tipping customary in Qatar ? What is considered "reasonable" amount to tip : 1. The guy that pushes the shopping trolley for you 2. The person that washes your car 3. The tea boy that makes coffee for you in the office 4. The waiters at 5-star restaurants 5. The petrol pump attendants etc

\vspace*{-1mm}
\textbf{Relevant} 

\item[$q'_2$] Tipping Beauty Salon. What do you think how much i should tip the stuff in a beauty salon for manicure/pedicure; massage or haircut?? Any idea what is required in Qatar?

\vspace*{-1mm}
\textbf{Relevant} 

$\ldots$

\item[$q'_9$] Business Meeting? Guys; I'm just enquiring about what one should wear to business meetings in Doha? Are there certain things a man should or shouldn't wear (Serious replys only - not like A man shouldn't wear a dress)!!!! Thanks - Gino

\vspace*{-1mm}
\textbf{Irrelevant} 

\item[$q'_{10}$] what to do? I see this man every morning; cleaning the road. I want to give him some money…(not any big amount)but I feel odd to offer money to a person who is not asking for it. I am confused; I kept the money handy in the car.... because of the traffic the car moves very slowly in that area; I can give it to him easily…..but am not able to do it for the past 4 days; and I feel so bad about it. If I see him tomorrow; What to do?

\vspace*{-1mm}
\textbf{Irrelevant} 
\vspace*{-3mm}
\end{description}
}
\caption{\label{fig:example} An input question and some of the potentially relevant questions retrieved for it.} 
\end{figure}

\section{Adversarial Training for Cross-Language Problems} 
\label{sec:model}

We demonstrate our approach for cross-language representation learning with adversarial training on a cross-lingual extension of the \emph{question--question similarity reranking} subtask of SemEval-2016 Task~3 on community Question Answering.

An example 
for the monolingual task is shown in Figure~\ref{fig:example}. We can see an original English input question $q$ and a list of several potentially similar questions $q_i^{\prime}$ from the Qatar Living\footnote{\url{http://www.qatarliving.com/forum}} forum, retrieved by a search engine. The original question (also referred to as a new question) asks about how to tip in Qatar. Question $q_1^{\prime}$ is relevant with respect to it as it asks the same thing, and so is  $q_2^{\prime}$, which asks how much one should tip in a specific situation. However, $q_9^{\prime}$ and $q_{10}^{\prime}$ are irrelevant: the former asks about what to wear at business meetings, and the latter asks about how to tip a kind of person who does not normally receive tips.


\noindent In our case, the input question $q$ is in a different language (Arabic) than the language of the retrieved questions (English).
The goal is to rerank a set of $K$ retrieved questions $\{q'_{k}\}_{k=1}^{K}$ written in a source language (e.g., English) according to their similarity with respect to an input user question $q$ that comes in another (target) language, e.g.,~Arabic. For simplicity, henceforth we will use Arabic as target and English as source. However, in principle, our method generalizes to any source-target language pair.

\subsection{Unsupervised Language Adaptation}

We approach the problem as a classification task, where given a 
question pair $(q, q')$, the goal is to decide whether the retrieved question $q'$ is \emph{similar} (i.e., relevant) to $q$ or not. Let $c\in\{0,1\}$ denote the class label: 1 for similar, and 0 for not similar. We use the posterior probability $p(c = 1|q,q',\theta)$ as a 
score for ranking all retrieved questions by similarity, where $\theta$ are 
the model parameters.

More formally, let $\Rs_n = \{q'_{n,k}\}_{k=1}^{K}$ denote the set of $K$ retrieved questions for a new question $q_n$. Note that the questions in $\Rs_n$ are always in English. We consider a training scenario where we have labeled examples $\Ds_S = \{q_n, q'_{n,k}, c_{n,k} \}_{n=1}^{N}$ for English $q_n$, but we only have unlabeled examples $\Ds_T = \{q_n, q'_{n,k}\}_{n=N+1}^{M}$ for Arabic $q_n$, with $c_{n,k}$ denoting the class label for the pair $(q_n, q'_{n,k})$. We want to train a cross-language model that can classify any test example $\{q_n,  q'_{n,k}\}$, where $q_n$ is in Arabic. This scenario is of practical importance, e.g., when an Arabic speaker wants to query the system in Arabic, 
and the database of related information is only in English. 
Here, we adapt the idea for adversarial training for domain adaptation as proposed by \newcite{Ganin:2016:DTN:2946645.2946704}. 

\begin{figure*}[t]
\centering
\hspace*{-1mm}\includegraphics[width=0.65\linewidth]{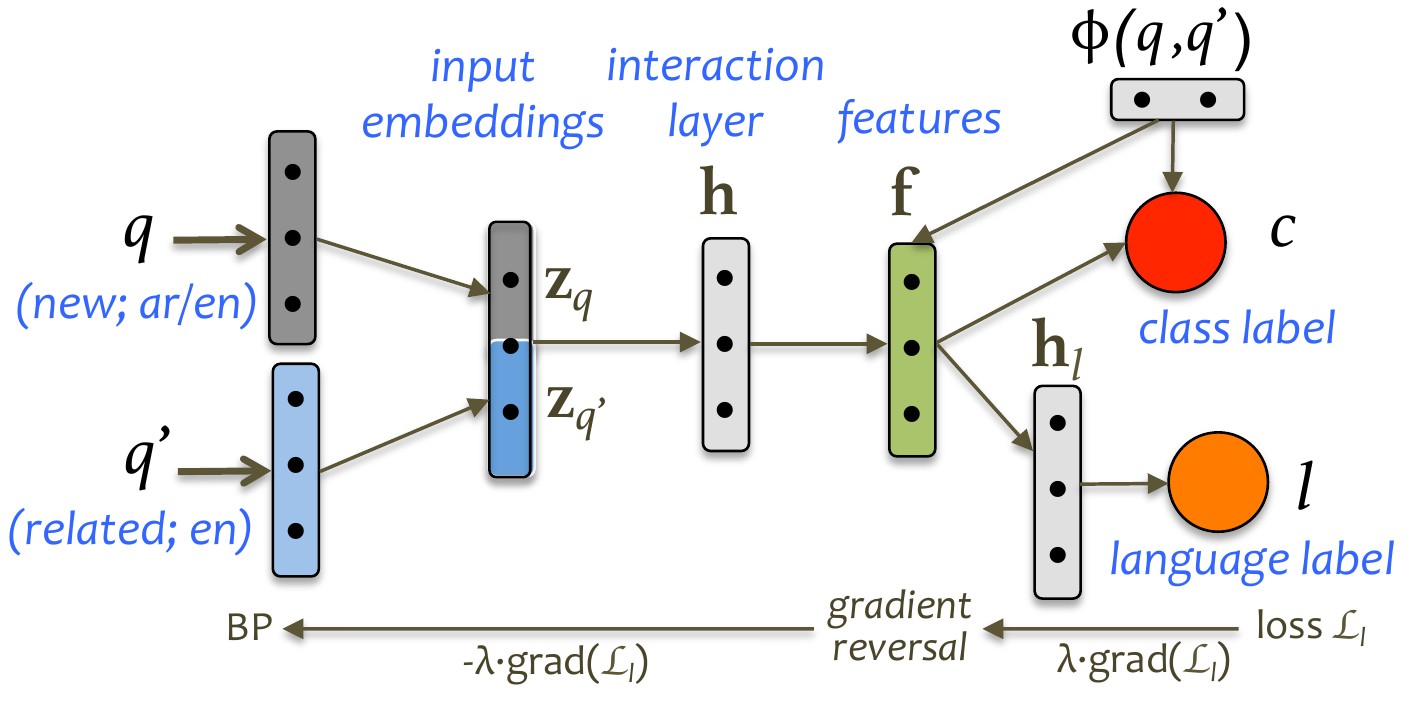}\caption{\label{f-clann} Architecture of CLANN for the question to question similarity problem in cQA.}
\end{figure*}

Figure \ref{f-clann} shows the architecture of our cross-language adversarial neural network (CLANN) model. The input to the network is a pair $(q, q')$, which is first mapped to fixed-length vectors $(\mathbf{z}_{q}, \mathbf{z}_{q'})$. 
To generate these word embeddings, one can use existing tools such as \emph{word2vec} \cite{Mikolov:2013} and monolingual data from the respective languages. Alternatively, one can use cross-language word embeddings, e.g., trained using the \emph{bivec} model \cite{luong-pham-manning:2015:VSM-NLP}. The latter can yield better initialization, which could be potentially crucial when the labeled data is too small to train the input representations with the end-to-end system.

\noindent The network then models the interactions between the input embeddings by passing them through two non-linear hidden layers, $\mathbf{h}$ and $\mathbf{f}$.  Additionally, the network considers \emph{pairwise} features $\phi(q,q')$ that go directly to the output layer, and also through the second hidden layer. 

The following equations describe the transformations through the hidden layers: 
\vspace*{-2mm}

\begin{eqnarray}
\mathbf{h} &=&  g(U [\mathbf{z}_{q};\mathbf{z}_{q'}]) \\
\mathbf{f} &=&  g(V [\mathbf{h};\phi(q, q')]) \label{eq:f}
\end{eqnarray}


\noindent where $[.;.]$ denotes concatenation of two column vectors, $U$ and $V$ are the weight matrices in the first and in the second hidden layer, and $g$ is a nonlinear activation function; we use rectified linear units or ReLU \cite{icml2010_NairH10}.  

The pairwise features $\phi(q,q')$ encode different types of similarity between $q$ and $q'$, and task-specific properties that we describe later in Section~\ref{sec:setting}. In our earlier work \cite{Giovanni2017_sigir}, we found it beneficial to use them directly to the output layer as well as through a hidden-layer transformation. The non-linear transformation allows us to learn high-level abstract features from the raw similarity measures, while the adversarial training, as we describe below, will make these abstract features language-invariant. 



The output layer computes a sigmoid: 
\vspace*{-2mm}

\begin{eqnarray}
\hat{c}_{\theta} = p(c = 1|\mathbf{f}, \mathbf{w}) =  \sigm (\mathbf{w}^T [\mathbf{f};\phi(q, q')])
\end{eqnarray}


\noindent where $\mathbf{w}$ are the output layer weights. 

We train the network by minimizing the negative log-probability of the gold labels: 


\begin{eqnarray}
\Ls_c(\theta) =  - c \log \hat{c}_{\theta} - (1-c) \log \left(1- \hat{c}_{\theta} \right)
\end{eqnarray}

\noindent The network described so far learns the abstract features through multiple hidden layers that are discriminative for the classification task, i.e., \emph{similar} vs. \emph{non-similar}. However, our goal is also to make these features invariant across languages. To this end, we put a language discriminator, another neural network that takes the internal representation of the network $\mathbf{f}$ (see Equation \ref{eq:f}) as input, and tries to discriminate between \emph{English} and \emph{Arabic} inputs --- in our case, whether the input comes from $\Ds_S$ or from $\Ds_T$. 

The language discriminator is again defined by a sigmoid function: 

\begin{eqnarray}
\hat{l}_{\omega} = p(l = 1|\mathbf{f}, \omega) = \sigm (\mathbf{w}_l^T \mathbf{h}_l)
\end{eqnarray}


\noindent where $l\in\{0,1\}$ denotes the language of $q$ (1 for English, and 0 for Arabic), $\mathbf{w}_l$ are the final layer weights of the discriminator, and $\mathbf{h}_l = g(U_l \mathbf{f})$ defines the hidden layer of the discriminator with $U_l$ being the layer weights and $g$ being the ReLU activations. 

We use the negative log-probability as the discrimination loss: 

\begin{eqnarray}
\Ls_l(\omega) =  - l \log \hat{l}_{\omega} - (1-l) \log \left(1- \hat{l}_{\omega} \right) \label{eq:dis-loss}
\end{eqnarray}

The overall training objective of the composite model can be written as follows:


\small
\begin{eqnarray}
\Ls(\theta,\omega) =  
\sum_{n=1}^N \Ls_c^n (\theta) - 
\lambda \Big[ \sum_{n=1}^N \Ls_l^n(\omega) + \sum_{n=N+1}^M \Ls_l^n(\omega) \Big]  \label{loss}
\end{eqnarray}
\normalsize

\noindent where $\theta = \{U,V,\mathbf{w}\}$, $\omega = \{U,V,\mathbf{w},U_l,\mathbf{w}_l\}$, and the hyper-parameter $\lambda$ controls the relative strength of the two networks. 

\noindent In training, we look for parameter values that satisfy a min-max optimization criterion as follows:

\begin{eqnarray}
\theta^* = \argmin_{U,V,\mathbf{w}} \max_{U_l,\mathbf{w}_l} \Ls (U,V,\mathbf{w}, U_l,\mathbf{w}_l) 
\end{eqnarray}

\noindent which involves a maximization (gradient ascent) with respect to $\{U_l,\mathbf{w}_l\}$ and a minimization (gradient descent) with respect to $\{U,V,\mathbf{w}\}$. Note that maximizing $\Ls (U,V,\mathbf{w}, U_l,\mathbf{w}_l) $ with respect to $\{U_l,\mathbf{w}_l\}$ is equivalent to minimizing the discriminator loss $\Ls_l(\omega)$ in Equation (\ref{eq:dis-loss}), which aims to improve the discrimination accuracy.  In other words, when put together, the updates of the shared parameters $\{U,V,\mathbf{w}\}$ for the two classifiers work adversarially with respect to each other. 

In our gradient descent training, the above min-max optimization is performed by reversing the gradients of the language discrimination loss $\Ls_l(\omega)$, when they are backpropagated to the shared layers. As shown in Figure \ref{f-clann}, the gradient reversal is applied to layer $\mathbf{f}$ and also to the layers that come before it.

Our optimization setup is related to the training method of Generative Adversarial Networks or GANs \cite{Goodfellow_14_GAN}, where the goal is to build deep generative models that can generate realistic images. The discriminator in GANs tries to distinguish real images from model-generated images, and thus the training attempts to minimize the discrepancy between the two image distributions, i.e., \emph{empirical} as in the training data vs. \emph{model-based} as produced by the generator. When backpropagating to the generator network, they consider a slight variation of the reverse gradients with respect to the discriminator loss. In particular, if $\rho$ is the discriminator probability, instead of reversing the gradients of $\log (1- \rho)$, they use the gradients of $\log \rho$. Reversing the gradient is a different way to achieve the same goal.



\paragraph{Training.}

Algorithm \ref{alg:training} shows pseudocode for the algorithm we use to train our model, which is based on stochastic gradient descent (SDG). We first initialize the model parameters by using samples from glorot-uniform distribution \cite{GlorotAISTATS2010}.
We then form minibatches of size $b$ by randomly sampling ${b}/{2}$ labeled examples from $\Ds_S$ and ${b}/{2}$ unlabeled examples from $\Ds_T$. For the labeled instances, both $\Ls_c(\theta)$ and  $\Ls_l(\omega)$ losses are active, while only the $\Ls_l(\omega)$ loss is active for the unlabeled instances. 

\small
\begin{algorithm}[t!]
\SetKwInOut{Input}{Input}\SetKwInOut{Output}{Output}
\SetAlgoNoLine
\SetNlSkip{0em}
\Input{data $\Ds_S$, $\Ds_T$, batch size $b$}
\Output{learned model parameters  $\{U,V,\mathbf{w}, U_l, \mathbf{w}_l \} $}
1. Initialize model parameters; \\
2. \Repeat {convergence}{ 
        
		\Na Randomly sample $\frac{b}{2}$ labeled examples from $\Ds_S$ \\
		\Nb Randomly Sample $\frac{b}{2}$ unlabeled examples from $\Ds_T$ \\
		\Nc Compute $\Ls_c(\theta)$ and $\Ls_l(\omega)$ \\
		\Nd Take a gradient step for $\frac{2}{b} \nabla_{\theta} \Ls_c(\theta) $  \\
		\Ne Take a gradient step for $\frac{2\lambda}{b} \nabla_{U_l,\mathbf{w}_l} \Ls_l(\omega)$  \\
		\tcp{Gradient reversal}
		\Nf Take a gradient step for $- \frac{2\lambda}{b} \nabla_{\theta} \Ls_l(\omega) $  \\

   }
\caption{Model Training with SGD}
\label{alg:training}
\end{algorithm}

\normalsize

\noindent As mentioned above, the main challenge in adversarial training is to balance the two components of the network. If one component becomes smarter, its loss to the shared layer becomes useless, and the training fails to converge \cite{ArjovskyCB17}. Equivalently, if one component gets weaker, its loss overwhelms that of the other, causing training to fail. In our experiments, the language discriminator was weaker. This could be due to the use of \emph{cross-language} word embeddings to generate input embedding representations for $q$ and $q'$.
To balance the two components, we would want the error signals from the discriminator to be fairly weak initially, with full power unleashed only as the classification errors start to dominate. We follow the weighting schedule proposed by \newcite [p. 21]{Ganin:2016:DTN:2946645.2946704}, who initialize $\lambda$ to $0$, and then change it gradually to $1$ as training progresses. I.e., we start training the task classifier first, and we gradually add the discriminator's loss.




\subsection{Semi-supervised Extension}
\label{subsec:semisup-model}

Above we considered an unsupervised adaptation scenario, where we did not have any labeled instance for the target language, i.e., when the new question $q_n$ is in Arabic. However, our method can be easily generalized to a semi-supervised setting, where we have access to some labeled instances in the target language, $\Ds_{T^*} = \{q_n, \Rs_n, c_n\}_{n=M+1}^{L}$. In this case, each minibatch during training is formed by labeled instances from both $\Ds_S$ and $\Ds_{T^*}$, and unlabeled instances from $\Ds_T$.  





\begin{table*}[htb]
\centering
\resizebox{0.95\linewidth}{!}{%
\begin{tabular}{lccccccc}
\hline
System& Input & Discrim. & Target & Hyperparam. (\emph{b}, \emph{d}, $\mathbf{h}$, $\mathbf{f}$, $l_2$) & MAP &	MRR & AvgRec \\\hline
FNN   & en & -- & ar & 8, 0.2, 10, 100, 0.03 & 75.28 & 84.26 & 89.48 \\
CLANN & en & en vs. ar' & ar & 8, 0.2, 15, 100, 0.02 & \bf 76.64 & \bf 84.52 & \bf 90.92\\\hline
FNN   & ar & -- & en & 8, 0.4, 20, 125, 0.03 & 75.32 &  84.17 & 89.26 \\
CLANN &	ar & ar vs. en' & en & 8, 0.4, 15, 75, 0.02 & \bf 76.70 & \bf 84.52 & \bf 90.61\\\hline
\end{tabular}
}
\caption{\label{t-results} Performance on the test set for our cross-language systems, with and without adversarial adaptation (CLANN and FNN, respectively), and for both language directions (en-ar and ar-en). The prime notation under the \emph{Discrim.} column represents using a counterpart from the unlabeled data.}
\end{table*}

\section{Experimental Setting}
\label{sec:setting}



In this section, we describe the datasets we used, the 
generation of the input embeddings, the nature of the pairwise features, and the general training setup of our model.

\subsection{Datasets}

SemEval-2016 Task 3~\cite{nakov-EtAl:2016:SemEval}, provides 267 input questions for training, 50 for development, and 70 for testing, and ten times as many potentially related questions retrieved by an IR engine for each input question: 2,670, 500, and 700, respectively.
%
%
Based on this data, we simulated a \emph{cross-language setup} for question-question similarity reranking. 
We first got the 387 original train+dev+test questions translated into Arabic by professional translators.
Then, we used these Arabic questions as an input with the goal to rerank the ten related English questions.
%
As an example, this is the Arabic translation of the original English question from Figure~\ref{fig:example}:

\hfill \<هل تعطون الاكراميات؟ ماذا تفعلون بهذا الشأن؛>

\hfill \<اذا كانت الاجابة نعم ، ما هو قوة ما تتجنبونه؟>

\hfill \<في بلادنا ، نترك من 15 إلى 20 بالمئة.>
\smallskip

We further collected 221 additional original questions and 1,863 related questions as unlabeled data, and we got the 221 English questions translated to Arabic.\footnote{Our cross-language dataset and code are available at \url{https://github.com/qcri/CLANN}}

\smallskip

\subsection{Cross-language Embeddings}
\label{subsec:xl-embeddings}

We used the TED \cite{ABDELALI14.877} and the OPUS parallel Arabic--English bi-texts~\cite{Tiedemann:12} to extract a bilingual dictionary, and to learn cross-language embeddings. We chose these bi-texts as they are conversational (TED talks and movie subtitles, respectively), and thus informal, which is close to the style of our community question answering forum.

\noindent We trained Arabic-English cross-language word embeddings from the concatenation of these bi-texts using \emph{bivec}~\cite{luong-pham-manning:2015:VSM-NLP}, a bilingual extension of \emph{word2vec}, which has achieved excellent results on semantic tasks close to ours~\cite{upadhyay-EtAl:2016:P16-1}. In particular, we trained 200-dimensional vectors using the parameters described in~\cite{upadhyay-EtAl:2016:P16-1}, with a context window of size 5 and iterating for 5 epochs. 
We then compute the representation for a question by averaging the embedding vectors of the words it contains. Using these cross-language embeddings allows us to compare directly representations of an Arabic or an English input question $q$ to English potentially related questions $q'_i$.



\subsection{Pairwise Features}
In addition to the embeddings, we also used some pairwise features that model the similarity or some other relation between the input question and the potentially related questions.\footnote{This required translating the Arabic input question to English. For this, we used an in-house Arabic--English phrase-based statistical machine translation system, trained on the TED and on the OPUS bi-texts; for language modeling, it also used the English Gigaword corpus.}
These features were proposed in the previous literature for the question--question similarity problem, and they are necessary to obtain state-of-the-art results.

In particular, we calculated the similarity between the two questions using 
machine translation evaluation metrics, as suggested in \cite{ACL2016:MTE-NN-cQA}. In particular, we used 
\textsc{Bleu} \cite{Papineni:Roukos:Ward:Zhu:2002};
\textsc{NIST} \cite{Doddington:2002:AEM};
\textsc{TER} v0.7.25 \cite{Snover06astudy};
\textsc{Meteor} v1.4 \cite{Lavie:2009:MMA} with paraphrases;
Unigram ~\textsc{Precision}; 
Unigram ~\textsc{Recall}.
We also used features that model various components of \textsc{Bleu}, as proposed in \cite{guzman-EtAl:2015:ACL-IJCNLP}:
$n$-gram precisions,
$n$-gram matches,
total number of $n$-grams ($n$=1,2,3,4),
hypothesis and reference length, 
length ratio,
and brevity penalty.

\noindent We further used as features the cosine similarity between question embeddings. In particular, we used \Ni 300-dimensional pre-trained Google News embeddings from \cite{Mikolov:2013},
\Nii 100-dimensional embeddings trained on the entire Qatar Living forum \cite{SemEval2016:task3:SemanticZ}, and \Niii 25-dimensional Stanford neural parser embeddings~\cite{socher-EtAl:2013:ACL2013}. The latter are produced by the parser internally, as a by-product.

Furthermore, we computed various task-specific features, most of them introduced in the 2015 edition of the SemEval task
by \newcite{nicosia-EtAl:2015:SemEval}. This includes some question-level features:
(1)~number of URLs/images/emails/phone numbers;
(2)~number of tokens/sentences;
(3)~average number of tokens;
(4)~type/token ratio;
(5)~number of nouns/verbs/adjectives/adverbs/ pronouns;
(6)~number of positive/negative smileys;
(7)~number of single/double/ triple exclamation/interrogation symbols;
(8)~number of interrogative sentences (based on parsing);
(9)~number of words that are not in \textsc{word2vec}'s Google News vocabulary.
Also, some question-question pair features:
(10)~count ratio in terms of sentences/tokens/nouns/verbs/ adjectives/adverbs/pronouns;
(11)~count ratio of words that are not in \textsc{word2vec}'s Google News vocabulary.
Finally, we also have one meta feature:
(12)~reciprocal rank of the related question in the list of related questions.


\subsection{Model settings} 
\label{subsec:clann-setting}

We trained our CLANN model by optimizing the objective in Equation~(\ref{loss}) using ADAM \cite{KingmaB14} with default parameters.  For this, we used up to 200 epochs. In order to avoid overfitting, we used dropout \cite{Srivastava14a} of hidden units, $l_2$ regularization on weights, and \emph{early stopping} by observing MAP on the development dataset ---if MAP did not increase for $15$ consecutive epochs, we exited with the best model recorded so far. We optimized the values of the hyper-parameters using grid search:
for minibatch (\emph{b}) size in $\{8, 12, 16\}$, for dropout (\emph{d}) rate in $\{0.2, 0.3, 0.4, 0.5\}$, for $\mathbf{h}$ layer size in $\{10, 15, 20\}$, for $\mathbf{f}$ layer size in $\{75, 100, 125\}$, and for $l_2$ strength in $\{0.01, 0.02, 0.03\}$.
The fifth column in Table~\ref{t-results} shows the optimal hyper-parameter setting for the different models. Finally, we used the best model as found on the development dataset for the final evaluation on the test dataset.

\section{Evaluation Results}
\label{sec:eval}


Below we present the experimental results for the unsupervised and semi-supervised language adaptation settings. We compare our cross-language adversarial 
network (CLANN) to a feed forward neural network (FNN) that has no adversarial part.

\subsection{Unsupervised Adaptation Experiments}

Table~\ref{t-results} shows the main results for our cross-language adaptation experiments. 
Rows 1-2 present the results when the target language is Arabic and the system is trained with English input. Rows 3-4 show the reverse case, i.e.,~adaptation into English when training on Arabic. 
\emph{FNN} stands for \emph{feed-forward neural network}, and it is the upper layer in Figure~\ref{f-clann}, excluding the language discriminator. \emph{CLANN} is the full \emph{cross-language adversarial neural network}, training the discriminator with English inputs paired with random Arabic related questions from the unlabeled dataset. We show three ranking-oriented evaluation measures that are standard in the field of Information Retrieval: mean average precision (MAP), mean reciprocal rank (MRR), and average recall (AvgRec). We
computed them using the official scorer from SemEval-2016 Task 3.\footnote{\url{http://alt.qcri.org/semeval2016/task3/}}
Similarly to that task, we consider Mean Average Precision (MAP) as the main evaluation metric. The table also presents, for reproducibility, the values of the neural network hyper-parameters after tuning (in the fifth column). 


\begin{table}[t!]
\centering
\resizebox{0.95\linewidth}{!}{%
\begin{tabular}{rlccc}
\hline
& System		  & MAP &	MRR & AvgRec \\\hline
\multicolumn{5}{l}{\bf Monolingual (English) from SemEval-2016} \\
1. & IR rank	& \emph{74.75} & \emph{83.79} & \emph{88.30} \\
2. & UH-PRHLT (1st)   & 76.70 & 83.02 & 90.31 \\
3. & ConvKN (2nd)    & 76.02 & 84.64 & 90.70 \\
\multicolumn{5}{l}{\bf Cross-language (Arabic into English)} \\
4. & CLANN      & 76.70 & 84.52 & 90.61\\\hline
\end{tabular}
}
\caption{\label{t-sota} Comparison of our 
cross-language approach (CLANN)
to the best results at SemEval-2016 Task 3, subtask B.}
\end{table}

\begin{figure*}[t!]
\centering
\begin{tabular}{cc}
\hspace*{-1mm}\includegraphics[width=0.49\linewidth]{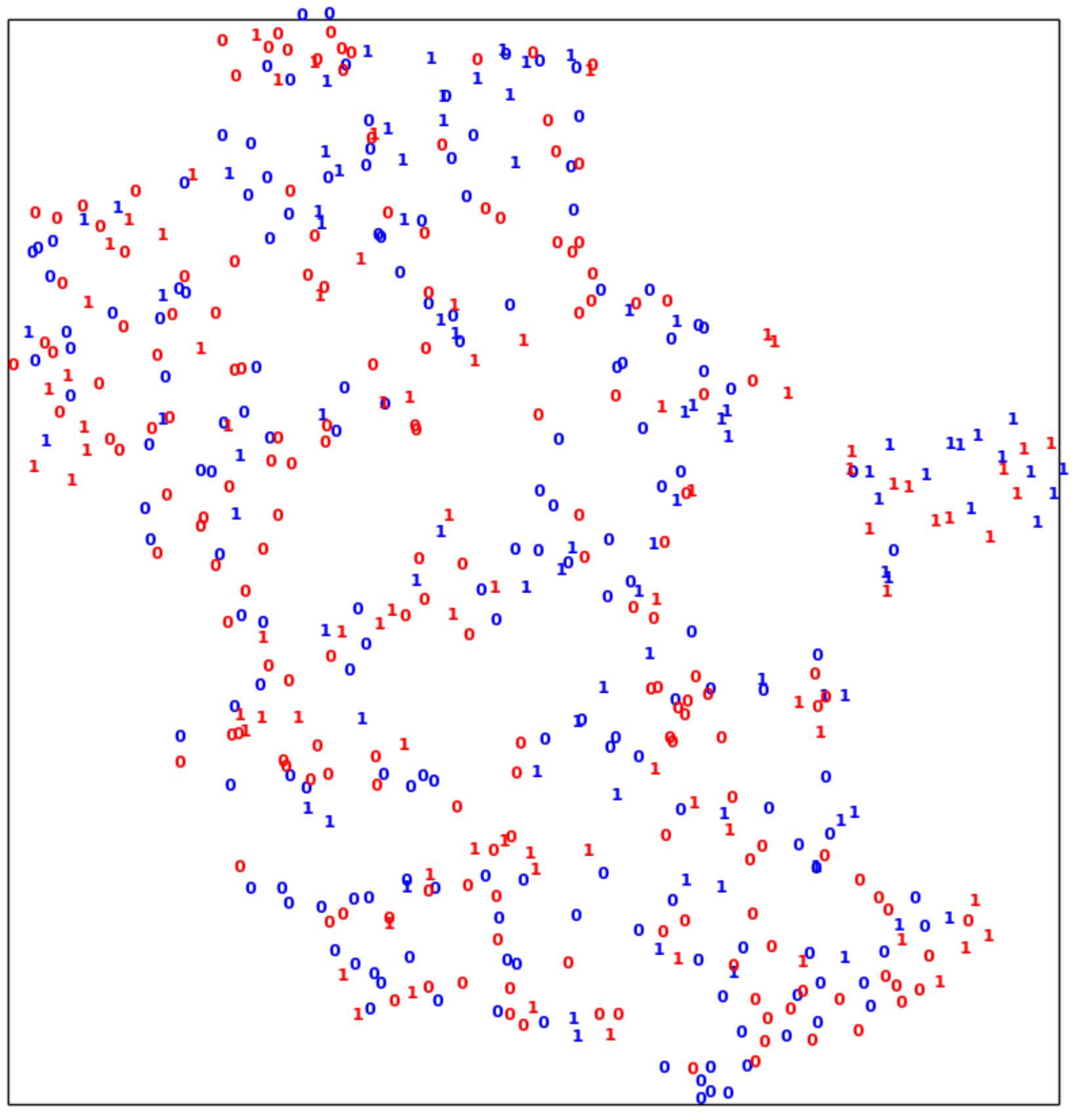} &
\hspace*{-3mm}\includegraphics[width=0.49\linewidth]{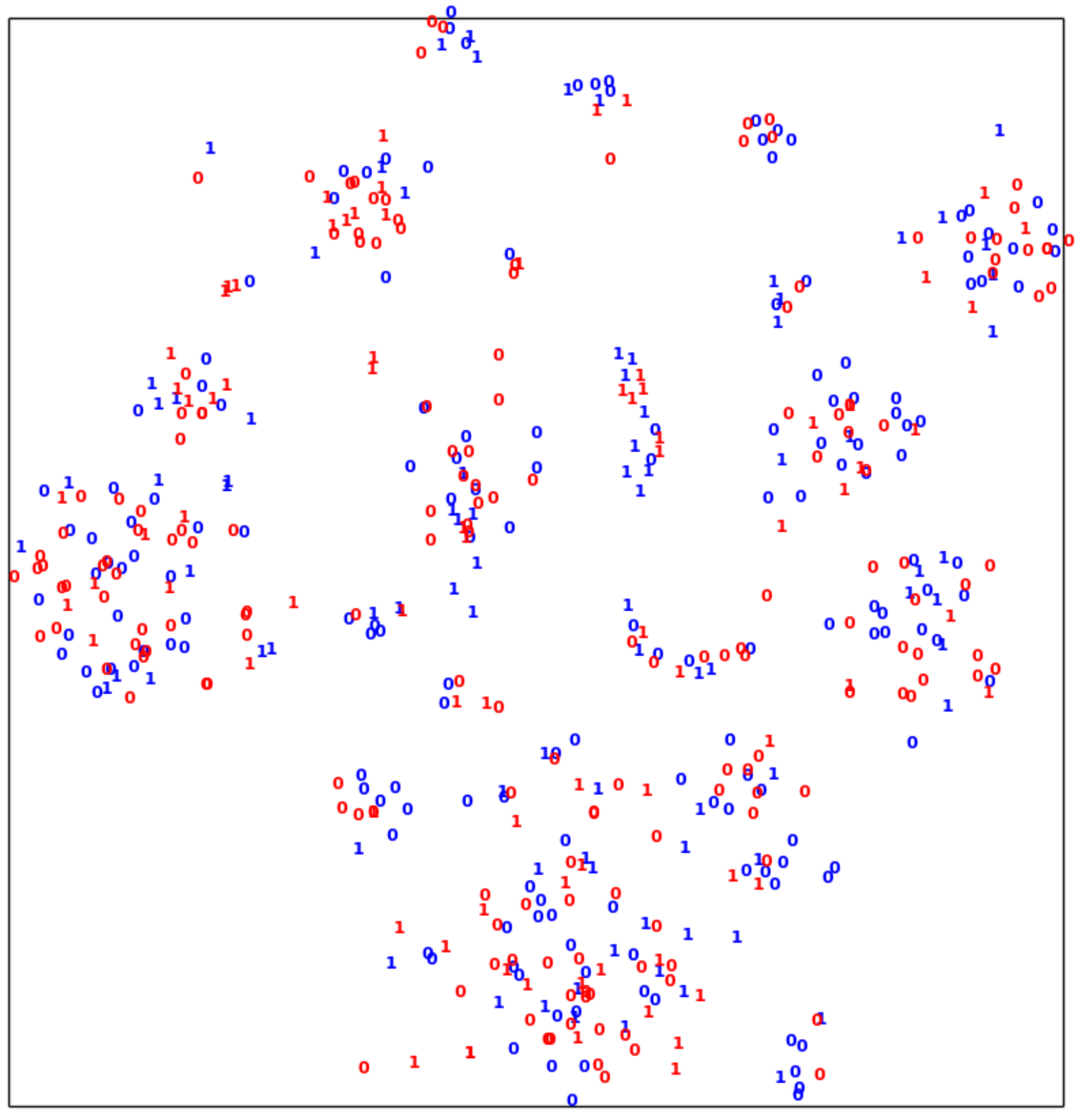} \\
\end{tabular}
\caption{\label{t-plots} Scatter plots showing Arabic and English test examples, after training the adversarial network. Arabic is shown in blue, and English is in red. 0-1 are the class labels. Left: ar$\rightarrow$en, right: en$\rightarrow$ar.}
%
\end{figure*}

We can see that the MAP score for FNN with Arabic target is 75.28. When doing the adversarial adaptation with the unlabeled Arabic examples (CLANN), the MAP score is boosted to 76.64 (+1.36 points). Going in the reverse direction, with English as the target, yields very comparable results: MAP goes from 75.32 to 76.70 (+1.38). 

\noindent To put these results into perspective, Table~\ref{t-sota} shows the results for the top-2 best-performing systems from SemEval-2016 Task 3, which used a monolingual English setting.
We can see that our FNN approach based on cross-language input embeddings is already not far from the best systems. 
Yet, when we consider the full adversarial network, in any of the two directions, we get performance that is on par with the best, in all metrics.


We conclude that the adversarial component in the network does the expected job, and improves the performance 
by focusing the language-independent features in the representation layer.
The scatter plots in Figure~\ref{t-plots} are computed by projecting the representation layer vectors of the first 500 test examples into two dimensions using t-SNE visualization~\cite{t-SNE}. The first 250 are taken with Arabic input (blue), the second 250 are taken with English input (red). 0-1 are the class labels (similar vs. non-similar). The top plot corresponds to CLANN training with English and adapting with Arabic examples, while the second one covers the opposite direction. The plots look as expected.
CLANN really mixes the blue and the red examples, as the adversarial part of the network pushes for learning shared abstract features that are language-insensitive. At the same time, the points form clusters with clear majorities of 0s or 1s, as the supervised part of the network learns how to classify them in these classes. 













\begin{table*}[tbh]
\centering
\resizebox{0.95\linewidth}{!}{%
\begin{tabular}{lcr@{}cccccc}
\hline
\multirow{2}{*}{System} & \multirow{2}{*}{Input} && \multirow{2}{*}{Discrim.} & \multirow{2}{*}{Target} & Hyperparam. & \multirow{2}{*}{MAP} &	\multirow{2}{*}{MRR} & \multirow{2}{*}{AvgRec} \\
& && & & (\emph{b}, \emph{d}, $\mathbf{h}$, $\mathbf{f}$, $l_2$) & & & \\\hline
FNN &	en && --- & ar & 8, 0.3, 10, 100, 0.03 & 74.69 & 83.79 & 88.16 \\
CLANN-unsup & en && en vs. ar' & ar & 12, 0.3, 15, 75, 0.02 & 75.93 & 84.15 & 89.63 \\
\multirow{2}{*}{CLANN-semisup} & \multirow{2}{*}{en+ar$^*$} & \multirow{2}{*}{$\Big\{$} & en vs. ar$^*$  & \multirow{2}{*}{ar} & \multirow{2}{*}{8, 0.4, 15, 75, 0.02} & \multirow{2}{*}{76.65} & \multirow{2}{*}{84.52} & \multirow{2}{*}{90.84} \\ 
& && en vs. ar' & & & & & \\
\hline
FNN &	ar && --- & en & 8, 0.2, 10, 75, 0.03& 75.38 & 84.05 & 89.12 \\
CLANN-unsup & ar && ar vs. en' & en & 12, 0.2, 15, 75, 0.03 & 75.89 & 84.29 & 89.54 \\
\multirow{2}{*}{CLANN-semisup} & \multirow{2}{*}{ar+en$^*$} &\multirow{2}{*}{$\Big\{$}& ar  vs. en$^*$  & \multirow{2}{*}{en} & \multirow{2}{*}{8, 0.2, 10, 75, 0.03} & \multirow{2}{*}{76.63} & \multirow{2}{*}{84.52} & \multirow{2}{*}{90.82} \\ 
& && ar vs. en' & & & & & \\
\hline
\end{tabular}
}
\caption{\label{t-semisup} Semi-supervised experiments, when training on half of the training dataset, and evaluating on the full testing dataset. Shown is the performance of our cross-language models, with and without adversarial adaptation (i.e., using CLANN and FNN, respectively), using the unsupervised and the semi-supervised settings, and for both language directions: English--Arabic and Arabic--English.  The prime notation in the \emph{Discrim.} column represents choosing a counterpart for the discriminator from the unlabeled data. The asterisks stand for choosing an unpaired labeled example from the other half of the training dataset.}
\end{table*}

\subsection{Semi-supervised Experiments}
\label{subsec:semisup}

We now study the semi-supervised scenario when we also have some labeled data from the target language, i.e., where the original question $q$ is in the target language.
This can be relevant in practical situations, as sometimes we might be able to annotate some data in the target language. It is also an exploration of training with data in multiple languages all together.

To simulate this scenario, we split the training set in two halves. We train with one half as the source language, and we use the other half with the target language as extra supervised data. At the same time, we also use the unlabeled examples as before. We introduced the semi-supervised model in subsection~\ref{subsec:semisup-model}, which is a straightforward adaptation of the CLANN model. 

Table~\ref{t-semisup} shows the main results of our cross-language semi-supervised experiments. The table is split into two blocks by source and target language (en-ar or ar-en). We also use the same notation as in Table~\ref{t-results}. The suffixes \emph{-unsup} and \emph{-semisup} indicate whether CLANN is trained in unsupervised mode (same as in Table~\ref{t-results}) or in semi-supervised mode. The language discriminator in this setting is trained to discriminate between labeled source and labeled target examples, and labeled source and unlabeled target examples. This is indicated in the \emph{Discrim.} column using asterisk and prime symbols, respectively.


\noindent There are several interesting observations that we can make about Table~\ref{t-semisup}. First, since here we are training with only 50\% of the original training data, both FNN and CLANN-unsup yield lower results compared to before, i.e., compared to Table~\ref{t-results}; this is to be expected. However, the unsupervised adaptation, i.e., using the CLANN-unsup model, still yields improvements over the FNN model by a sizable margin, according to all three evaluation measures. When we also train using the additional labeled examples in the target language, i.e., using the CLANN-semisup model, the results are boosted again to a final MAP score that is very similar to what we had obtained before with the full source-language training dataset. 
In the English into Arabic adaptation, the MAP score jumps from 74.69 to 76.65 (+1.96 points) when going from the FNN to the CLANN-semisup model, the MRR score goes from 83.79 to 84.52 (+0.73), and the AvgRec score is boosted from 88.16 to 90.84 (+2.68). The results in the opposite adaptation direction, i.e.,~from Arabic into English, follow a very similar pattern.

These results demonstrate the effectiveness and the flexibility of our general adversarial training framework within our CLANN architecture when applied to a cross-language setting for question-question similarity, 
taking advantage of the unlabeled examples in the target language (i.e., when using unsupervised adaptation) and also taking advantage of any labeled examples in the target language that we may have at our disposal (i.e., when using semi-supervised training with input examples in the two languages simultaneously).



\section{Conclusion}
\label{sec:conclusion}

We have studied the problem of cross-language adaptation for the task of question--question similarity reranking in community question answering,
when the input question can be either in English or in Arabic
with the objective to port a system trained on one input language to another input language 
given labeled data for the source language and only unlabeled data for the target language.
We used a discriminative adversarial neural network, which we trained to learn task-specific representations directly. This is novel in a cross-language setting, and we have shown that it works quite well.
The evaluation results have shown sizable improvements over a strong neural network model that uses simple projection with cross-language word embeddings.


In future work, we want to extend the present research in several 
directions. For example, we would like to start with monolingual word embeddings and to try to learn the shared cross-language representation directly as part of the end-to-end training of our neural network. We further plan to try LSTM and CNN for generating the initial representation of the input text (instead of simple averaging of word embeddings).
We also want to experiment with more than two languages at a time. Another interesting research direction we want to explore is to try to adapt our general CLANN framework to other tasks, e.g., to answer ranking in community Question Answering \cite{Joty:2016:NAACL,nakov-marquez-guzman:2016:EMNLP2016} in a cross-language setting, 
as well as to cross-language representation learning for words and sentences.

\section*{Acknowledgment}
This research was performed by the Arabic Language Technologies group at Qatar Computing Research Institute, HBKU\@, within the Interactive sYstems for Answer Search project ({\sc Iyas}).



\bibliography{references}
\bibliographystyle{acl_natbib}

\end{document}